\documentclass[letterpaper]{article}
\usepackage{aaai19}
\usepackage{times}
\usepackage{helvet}
\usepackage{courier}
\usepackage{graphicx}
\usepackage{epsfig} 
\usepackage{graphics} 
\usepackage{color}
\usepackage{subfigure}
\usepackage{amsmath}
\usepackage{amssymb}
\usepackage{amsfonts}
\usepackage{flafter}
\usepackage{booktabs}
\usepackage{blindtext}
\usepackage{color}
\usepackage{lmodern}
\usepackage{marginnote,url}
\usepackage[usenames,dvipsnames]{xcolor}

\definecolor{Red}{rgb}{0.7,0.0,0.0}
\definecolor{Green}{rgb}{0.0,0.7,0.0}
\definecolor{Blue}{rgb}{0.0,0.0,0.7}

\long\def\secret#1{\fbox{\textsl{Anonymized}}}

\definecolor{green1}{RGB}{51,204,0}
\definecolor{red1}{RGB}{255,51,0}
\definecolor{brown1}{RGB}{102,0,0}
\definecolor{pink1}{RGB}{255,51,255}
\definecolor{yellow1}{RGB}{255,204,0}
\definecolor{cyan1}{RGB}{0,255,255}
\definecolor{violet1}{RGB}{102,102,255}
\definecolor{purple1}{RGB}{153,0,153}
\definecolor{green2}{RGB}{0,102,102}
\definecolor{yellow2}{RGB}{255,153,0}
\definecolor{blue1}{RGB}{0,102,204}
\definecolor{seagreen1}{RGB}{0,255,255}
\definecolor{green3}{RGB}{51,51,0}
\definecolor{salmon}{RGB}{255,153,153}
\definecolor{green4}{RGB}{204,204,0}

\usepackage{soul}
\definecolor{reddish}{HTML}{FBB4AE}
\definecolor{blueish}{HTML}{B3CDE3}
\definecolor{magentish}{HTML}{FF00AA}
\definecolor{greenish}{HTML}{a1d99b}

\begin{document}

\title{Towards Automatic Clustering Analysis using Traces of Information Gain: The InfoGuide Method}
\author{Paulo Rocha\textsuperscript{\rm 1}, Diego Pinheiro\textsuperscript{\rm 2}, Martin Cadeiras\textsuperscript{\rm 2} and Carmelo Bastos-Filho\textsuperscript{\rm 1}\\
 \textsuperscript{\rm 1} Department of Computer Engineering, University of Pernambuco, Brazil\\
 {\tt \{phar, carmelofilho\}@poli.br}\\
 \textsuperscript{\rm 2} Department of Internal Medicine, University of California, Davis, US\\
 {\tt \{pinsilva, mcadeiras\}@ucdavis.edu}\\
 }
% \author{Authors are temporally omitted due to the requirements regarding the blind review process}
\maketitle

\begin{abstract}
Clustering analysis has become a ubiquitous information retrieval tool in a wide range of domains, but a more automatic framework is still lacking. 
Though internal metrics are the key players towards a successful retrieval of clusters, their effectiveness on real-world datasets remains not fully understood, mainly because of their unrealistic assumptions underlying datasets. 
We hypothesized that capturing {\it traces of information gain} between increasingly complex clustering retrievals---{\it InfoGuide}---enables an automatic clustering analysis with improved clustering retrievals.
We validated the {\it InfoGuide} hypothesis by capturing the traces of information gain using the Kolmogorov-Smirnov statistic and comparing the clusters retrieved by {\it InfoGuide} against those retrieved by other commonly used internal metrics in artificially-generated, benchmarks, and real-world datasets.
Our results suggested that {\it InfoGuide} can enable a more automatic clustering analysis and may be more suitable for retrieving clusters in real-world datasets displaying nontrivial statistical properties.
\end{abstract}
% \smallskip
% \noindent \textbf{Keywords}\\  Social Networks, Linguistics, Twitter, Human Development Index.

\section{Introduction}
\label{Introduction}

Clustering analysis has become ubiquitous in the retrieval of clusters from a plethora of datasets arising from a wide range of domains~\cite{Adolfsson:2019bv}, supporting the characterization of 
%heterogeneous diseases~\cite{Seymour:2019fd}, 
the development of personalized medical therapies~\cite{Bakir:2018cb}, 
the understanding of intricate social-economic factors~\cite{Mirowsky:2017fv}, and the development of healthcare ranking systems~\cite{Wallace:2019fd}. Since the creation of the first clustering algorithm in 1948, by the botanist and evolutionary biologist Thorvald Sørensen while studying biological taxonomy~\cite{sorensen1948method}, novel algorithms for clustering retrieval have been proposed~\cite{xu2015comprehensive}. However, a framework for automatic clustering analysis is still lacking ad even determining the optimal number of clusters to be retrieved remains a major methodological issue~\cite{Tibshirani:2001fj}. 

Given that ground truth labels are inherently absent, the clustering retrieval largely relies on internal metrics of clustering quality~\cite{arbelaitz2013extensive}. These metrics are idealized aspects of clustering quality defined a priori and often involve unrealistic assumptions about the datasets~\cite{Tibshirani:2001fj,rousseeuw1987silhouettes,calinski1974dendrite}. Nevertheless, clustering algorithms not only, directly or indirectly, retrieve clusters according to these internal metrics~\cite{macqueen1967some,ward1963}, but also have their clustering retrieval subsequently evaluated according to these same metrics. As a result, different internal metrics often disagree with each other regarding the quality of a specific clustering retrieval~\cite{Tibshirani:2001fj,xu2015comprehensive}. Another crucial issue relies on the fact that most of the metrics use distances and, sometimes, the distances in different attributes of the problem may have different meanings.

We hypothesized that capturing the {\it traces of information gain} between increasingly complex clustering retrievals---the {\it InfoGuide} method---can enable a more automatic clustering analysis. We validated the {\it InfoGuide} hypothesis by capturing the traces of information using the Kolmogorov–Smirnov statistic and comparing the clusters retrieved by {\it InfoGuide} against those retrieved by other commonly used internal metrics over artificially-generated, benchmarks, and real-world datasets. Our results suggest that {\it InfoGuide} may be more suitable to retrieve clusters in real-world datasets displaying nontrivial statistical properties. 

\section{Related Work}
\label{Related Work}

The application of a clustering algorithm $g$ over a dataset  $\mathcal{X}$ to retrieve $k$ groups, a clustering retrieval $C^{(k)}$, can be generally defined as a mapping $g_{k}: \mathcal{X}  \rightarrow C^{(k)}$  such that each data point $x \in \mathcal{X}$ is assigned to one of the $k$ clusters $c_i^{k} \in C^{(k)}$.  Each $c^{(k)}_i$ in $C^{(k)}$ represents a subgroup of the dataset $\mathcal{X}^{(N_i,F)}_i$ as following: 

\begin{equation}
\begin{aligned}
C^{(k)} = \{c^{(k)}_1, c^{(k)}_2, \dots, c^{(k)}_k\} \enspace , 
\end{aligned}
\label{Eq:C}
\end{equation}
in which $N_i$ is the number of data points in $c^{(k)}_i$.
%, and $N$ is equal to the sum of $N_i \in [1, k]$:

%
%\begin{equation}
%\begin{aligned}
%   c^{(k)}_i = \mathcal{X}^{(N_i,F)}_i
%\end{aligned}
%\label{Eq:c_i}
%\end{equation}

%Let $\mathcal{X}^{N,F}$ be a dataset with $N$ data points and $F$ features, and $g$  clustering algorithm, $g_{k}: \mathcal{X}^{(N,F)} \rightarrow C^{(k_{min})}$ is the mapping generated by the application of $g$ over $\mathcal{X}^{N,F}$,$$ that generates a set $C^{(k)}$ with $k$ groups from $X^{(N,F)}$ where $k \in [k_{min}, k_{max}]$:
%
% \begin{equation}
% \begin{aligned}
%     &g_{k_{min}}: X^{(N,F)} \rightarrow C^{(k_{min})} \\
%     &g_{k_{min + 1}}: X^{(N,F)} \rightarrow C^{(k_{min + 1})} \\
%     &\;\vdots \\
%     &g_{k_{max}}: X^{(N,F)} \rightarrow C^{(k_{max})}
% \end{aligned}
% \label{Eq:g}
% \end{equation}

Clustering algorithms can be classified into different categories according to their assumptions about clustering retrieval, namely, partitions, hierarchy, density, distribution, subspace, to name but a few~\cite{rodriguez2019clustering,xu2015comprehensive}. Despite the differences among categories, the main idea underlying clustering retrieval is that data points belonging to the same cluster should be similar to each other and dissimilar from data points belonging to other clusters~\cite{sorensen1948method}.

% A similarity measure needs to be established to assess the extent to which groups resemble the main idea of clustering. Such similarity is often proposed as a measure of the distance between data points. In simple approaches, the similarity between two data points depends on the distance between them~\cite{sorensen1948method}. The chosen distance (e.g., Euclidean, Manhattan, Mahalanobis) applied to the $g_k$ can generate a bias in the shape of the groups, resulting in different clustering retrievals for the same $\mathcal{X}$ and $k$.

% To assess the clustering retrievals, different cluster validity indexes, also called internal metrics, were established to proper evaluate the quality of the mapping $g_{k}: \mathcal{X}  \rightarrow C^{(k)}$, this process can be described as a mapping $m_{k}: C^{(k)}  \rightarrow q$ were $q$ represents a comparable scalar which enables the comparison between different numbers of $k$, resulting in the possibility of finding the optimal number of clusters $\hat{k}$ for ${g_k} \in [k_{min}, k_{max}]$.

In general, the similarity between two data points depends on their distance~\cite{sorensen1948method}. The chosen distance (e.g., Euclidean, Manhattan, Mahalanobis) applied to the $g_k$ can generate a bias in the shape of the groups, resulting in different clustering retrievals even  the same $\mathcal{X}$ and $k$. To proper evaluate the quality of the mapping $g_{k}: \mathcal{X}  \rightarrow C^{(k)}$, different internal metrics $m_{k}: C^{(k)}  \rightarrow q$ have been proposed, in which $q$ represents a comparable scalar enabling the comparison between different numbers of $k$ as well as the possibility of finding the optimal number of clusters $\hat{k}$ for ${g_k} \in [k_{min}, k_{max}]$.

In the extensive study of Arbelaitz et al, 30 internal metrics were evaluated in a wide range of datasets, demonstrating that most of the metrics simply determine the quality of clustering retrievals by applying primarily two criteria: the distance between points in the same cluster, described as {\it cohesion}, and the distance between different groups,  described as {\it separation}. These metrics have a bias that better evaluates a set of groups with both low cohesion and high separation, and can thus be defined as distance-based internal metrics~\cite{arbelaitz2013extensive}.

Conversely, an information theoretic measure of cluster separability was developed by Gokcay and Principe as a cost function to guide an optimization clustering algorithm~\cite{gokcay2002information}. The authors used both artificially generated and image segmentation datasets. Similarly, Faivishevsky and Goldberger proposed the clusters mutual information as the maximization objective to be used by a clustering algorithm~\cite{faivishevsky2010nonparametric}. The authors demonstrated that an entropy-based approach may be more suitable than a distance-based approach for clustering analysis over both artificially-generated and benchmark datasets.

In this paper, a information theoretic approach for clustering analysis was moved forward given the main challenges faced by distance-based metrics especially in real-world datasets with nontrivial distributions, in which concepts such as averages and distance-based similarities become unrealistic~\cite{gokcay2002information}. In this sense, we proposed the {\it InfoGuide} method for automatic clustering analysis in which an optimal clustering retrieval is based on the information gained between increasingly complex clustering retrievals.

\section{Methods}
\label{Methods}

\begin{figure*}[ht!]
	\centerline{\includegraphics[width=7.1in]{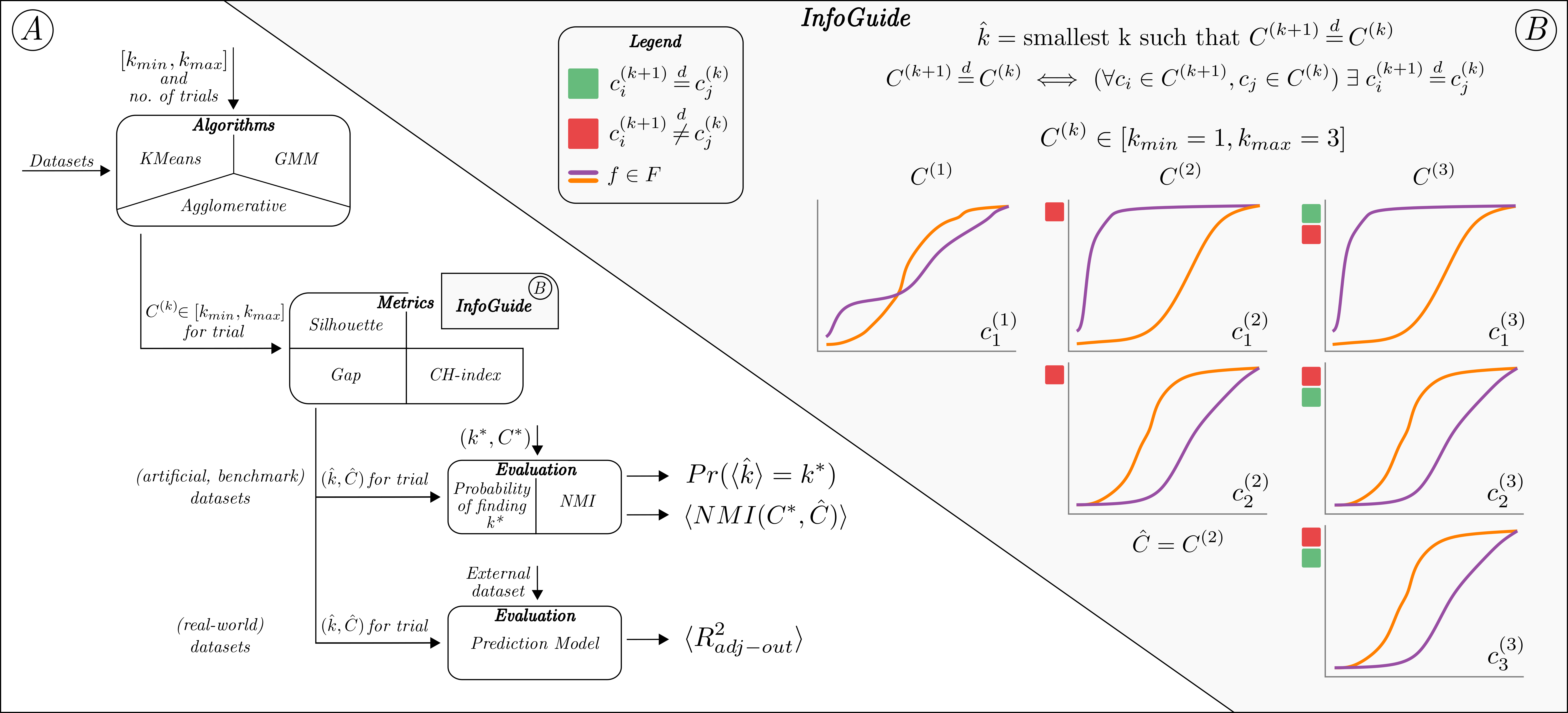}}
	\caption{(A) Clustering Analysis. For each dataset, a clustering retrieval is obtained by each clustering algorithm. Retrievals are subsequently evaluated by internal metrics, and an optimal clustering retrieval $\hat{C})$ is determined by each metric. When ground-truth is available, the probability of finding the true number of clusters $k^*$ and the normalized mutual information $NMI(C^*, \hat{C})$ are evaluated. In the absence of ground-truth, the goodness of fit of a prediction model (e.g., the out-sample adjusted $R^2$, $R^2_{adj-out}$) is evaluated when the clustering retrieval $\hat{C}$ is included as an additional predictor. (B) Illustration of the {\it InfoGuide} method for $k_{min} = 1$, $k_{max} = 3$, and $k^*=2$. The {\it InfoGuide} evaluates the equivalency between increasingly complex clustering retrievals $C^{(k)}$ and $C^{(k+1)}$. There is no information gain, for instance, when comparing $C^{(2)}$ and $C^{(3)}$  because for each cluster in $ C^{(3)}$ there is at least one equivalent cluster in $C^{(2)}$. As a result, the optimal number of clusters is $\hat{k}=2$.
	}
	\label{fig:method}
\end{figure*}

Clustering analysis involves the following elements: a dataset, a set of clustering algorithms, as well as internal and external metrics of clustering retrieval (\figurename~\ref{fig:method}, A). In this work, we proposed the {\it InfoGuide} method for automatic clustering analysis using traces of information gain (\figurename~\ref{fig:method}, B).

\subsection{InfoGuide---An Automatic Retrieval of Clusters using Traces of Information Gain}

% From that there is a function $g^{-1}_{k}$ which for each $l_i$ in the set $L_k$ can retrieve the respective group of data $X^{i}_{N_i, f}$ with $N_i$ data points and $f$ features:
% \begin{equation}
%     g^{-1}_k: l_i \rightarrow X^{i}_{N_i,f}
%     \label{Eq:g-1}
% \end{equation}

The challenge in clustering analysis is retrieving the highest number $\hat{k}$ of {\it meaningful} clusters as close to the optimal number $k^*$  of clusters as possible, avoiding both underfitting $\hat{k} < k^*$ and overfitting $\hat{k} > k^*$. 
%Retrieving $1$ or $N$ clusters are generally undesired trivial solutions. Often,  
The definition of a {\it meaningful} cluster not only depends on the specific internal metric used but also is affected by the specific clustering algorithm employed. 

Let $C^{(k)}$ and $C^{(k+1)}$ be the set of $k$ and $k+1$ increasingly complex clustering retrievals, respectively, the {\it InfoGuide} method retrieves the smallest number of clusters $\hat{k}$ as long as an increased information gain can be obtained between increasingly complex clustering retrievals as following:

\begin{equation}
\hat{k} = \textrm{smallest k such that } C^{(k+1)} \,{\buildrel d \over =}\, C^{(k)} \enspace ,
\label{Eq:selectk}
\end{equation}
in which the clustering retrieval $C^{(k+1)}$ is equivalent to $C^{(k)}$ according to the pairwise equivalencies between their individual clusters as following: 

\begin{equation}
\begin{aligned}
& C^{(k+1)} \,{\buildrel d \over =}\, C^{(k)} \iff \\
  &  (\forall c_i \in C^{(k+1)}, c_j \in C^{(k)}) \; \exists\; c^{(k+1)}_i \,{\buildrel d \over =}\, c^{(k)}_j \enspace ,
  \label{Eq:C=C}
\end{aligned}
\end{equation}
%C^{(k+1)} \,{\buildrel d \over =}\, C^{(k)} \iff \\ (\forall c_i \in C^{(k+1)}, c_j \in C^{(k)}) \; \exists\; c^{(k+1)}_i \,{\buildrel d \over =}\, c^{(k)}_j \enspace ,
%\end{aligned}
%\label{Eq:C=C}
%\end{equation}
in which individual clusters $c^{(k+1)}_i$ and  $c^{(k)}_j$ are as following:

\begin{equation}
\begin{aligned}
c^{(k+1)}_i \,{\buildrel d \over =}\,c^{(k)}_j \iff (\forall f \in F)  \; f_i \,{\buildrel d \over =}\, f_j \enspace ,
\end{aligned}
\label{Eq:f}
\end{equation}
in which the feature $f$ in $c^{(k+1)}_i$  and $c^{(k)}_i$ are equivalent in distribution.  Therefore, the {\it InfoGuide} method only considers that the clustering retrieval $C ^ {(k + 1)}$ increases the information gain relative to $C_ {k}$ when it retrieves novel clusters not already contained in $C_ {k}$. Otherwise, retrieving a higher number of clusters only results in a more complex model without information gain.  

In this work, the Kolmogorov-Smirnov $KS$ statistic was used to quantify the equivalency in distribution between features such that $f_i \,{\buildrel d \over =}\, f_j \equiv KS(f_i, f_j)$. Information gain is thus the statistical evidence that both features may not come from the same statistical distribution whenever the p-value of the $KS$ test is lower than the statistical significance $\alpha$ after using the Bonferroni correction for the $F \times (k+1) \times k$ multiple comparisons. The optimal $\hat{k} \in [k_{min}, k_{max}]$ is the highest $\hat{k}$ that can be obtained for a range of $\alpha_u \in (0, \alpha]$.

\subsection{Metrics}

The {\it InfoGuide} method was compared with three commonly used internal metrics that embrace the two main ideas underlying clustering analysis, namely, cohesion and separation. Let $\mathcal{X}^{N}$ be a dataset with $N$ data points and $C^{(k)}$ a clustering retrieval, these internal metrics use the following basic calculations of distance: between two data points, $(x_i - x_j)$, between a data point and the estimated value of a group, $(x_i - \langle c^{(k)}_i \rangle)$, and between the estimated values of a group and a dataset, $(\langle c^{(k)}_i \rangle - \langle C^{(k)} \rangle)$.

%\subsubsection{Silhouette}
The {\it Silhouette} chooses the optimal $\hat{k}$ by maximizing the average difference between the separation and cohesion as following~\cite{rousseeuw1987silhouettes}: 

\begin{equation}
SI = \frac{1}{N} \sum\limits_{i=1}^{N} \frac{b_i\; - \;a_i}{max(b_i\;,\;a_i)} \enspace ,
\label{Eq:silhouette}
\end{equation}
in which $a_i$ measures the cohesion of a data point $i$ as following:

\begin{equation}
a_i = \frac{1}{N_i - 1} \sum\limits_{j=1, j\neq i}^{N_i} (x_i - x_j) \enspace ,
\label{Eq:ai}
\end{equation}
and $b_i$ measures the separation of a data point $i$ to the other points belonging to nearest cluster as following: 

\begin{equation}
b_i = \min\limits_{1\leq l \leq k ,x_i \not\subset c^{(k)}_i} \bigg( \frac{1}{N_l} \sum\limits_{j=1}^{N_l} (x_i - x_j) \bigg) \enspace .
\label{Eq:bi} 
\end{equation}

%\subsubsection{Calinsk-Harabasz Index}
The {\it Calinsk-Harabasz (CH) Index} chooses the optimal $\hat{k}$ by maximizing the ratio between the Sum of Squares Within (SSW) and the Sum of Squares Between (SSB) as following~\cite{calinski1974dendrite}: 

\begin{equation}
CH = \frac{N - k}{k - 1} \cdot \frac{SSB}{SSW} \enspace ,
\label{Eq:ch-index}
\end{equation}
in which  $ SSW = \sum\limits_{i=1}^{k}\sum\limits_{j=1}^{N_i} (x_j - \langle c^{(k)}_i\rangle)^2$ is a measure of cohesion and $    SSB = \sum\limits_{i=1}^{k} N_i \cdot (\langle c^{(k)}_i \rangle - \langle C^{(k)}\rangle)^2$ is a measure of separation.  It is normalized by the number of data points $N$ and the number of groups $k$ to ensure a similar scale when comparing different numbers of groups.

%and separation, respectively\cite{calinski1974dendrite}. 
%On equation \ref{Eq:ch-index}

%\begin{equation}
%   
%\label{Eq:ssw}
%\end{equation}

%\begin{equation}
%    SSB = \sum\limits_{i=1}^{k} N_i \cdot (\langle c^{(k)}_i \rangle - \langle C^{(k)}\rangle)^2
%\label{Eq:ssb}
%\end{equation}

%\subsubsection{Gap Statistic}

The {\it Gap Statistic} maximizes the $SSW$ of a clustering retrieval from the actual dataset relative to what would be expected by a clustering retrieval from an uniformly distributed dataset $SSW_{random}$ as following~ \cite{Tibshirani:2001fj}:

\begin{equation}
Gap =  \mathbb{E}(\log{(SSW_{random})}) - \log{(SSW)} \enspace ,
\label{Eq:gap}
\end{equation}
such that greater the difference between random and actual cohesions, the higher the quality of the clustering retrieval. The optimal $\hat{k}$ is chosen as the smallest $k$ where $Gap(k) \geq Gap(k+1) - S_{k+1}$, and $S_{k+1}$ is the standard deviation of $\log{SSW_{random}}$.

\subsection{Experimental Setup}

The {\it InfoGuide} method was validated by comparing the quality of its clustering retrievals with those of other internal metrics over artificially-generated, benchmarks, and real-world datasets. Three commonly used clustering algorithms with distinct underlying approaches were used: K-Means~\cite{macqueen1967some}, Gaussian Mixture Model (GMM)~\cite{rasmussen2000infinite} and the Agglomerative Ward which is a Hierarchical Agglomerative with Ward's linkage~\cite{ward1963}. For each algorithms, $k \in [k_{min}, k_{max}]$  clusters were repeatedly retrieved $30$ times with $k_{min}=1$ and $k_{max}=11$. The optimal clustering retrieval $\hat{C}$ was obtained according to the {\it InfoGuide} as well as to the Silhouette, Calinsk-Harabasz Index, and Gap Statistic. A total of $7,920$ clustering retrievals $C^{(k)}$ were considered using $8$ datasets $\times$ $30$ trials $\times$ $3$ algorithms $\times$ $|[k_{min}, k_{max}]| = 11$ number of clusters. 

%In the final step each pair $(\hat{k}, \hat{C})$ were evaluated by external validation indexes. 

For artificially-generated and benchmark datasets, for which the ground truth $C^*$ are available, two evaluations were performed: the probability of finding the true $k^*$, $Pr(\langle \hat{k} \rangle = k^*)$, and the Normalized Mutual Information ($NMI(C^*, \hat{C})$) between the clusters retrieved $\hat{C}$ and ground-truth $C^*$. The probability of finding $k^*$ was quantified using the Wilson Score, which estimates the population proportion of a binomial distribution in which a success is encoded as $\hat{k} = k^*$. The $NMI$ quantifies the decrease in the entropy of $\hat{C}$ by knowing $C^*$. 

For real-world datasets, an external evaluation was performed by quantifying the goodness of fit of a prediction model when the optimal clustering retrieval $\hat{C}$ is included as an additional predictor. In this work, a Linear Regression was used, and the goal was to compare different metrics instead of obtaining the best prediction model. To control for model complexity and avoids overfitting, the adjusted $R^2$ out-sample, $R^2_{adj-out}$ was used. All of the code, datasets, and analysis are available on the Open Science Framework (OSF) repository of this project at \url{https://doi.org/10.17605/OSF.IO/ZQYNC}.

%T he second step consisted of the estimation of the optimal pair $(\hat{k}, \hat{C})$
%for the clustering retrievals of each combination algorithm-dataset-trial by the internal validation indexes. The three distance based metrics described above and the {\it InfoGuide} were used to analyse a total of $7,920$ sets of groups $C^{(k)}$. 

%performed experiment (\figurename~\ref{fig:method}, A) consisted OF three main steps: obtaining the clustering retrievals, estimating the optimal number of clusters $\hat{k}$ and, respectively, the set of clusters ${\hat{C}}$, and internally and externally evaluating the found pairs of $(\hat{k}, \hat{C})$.

%, representing the categories based on partition, distribution and hierarchy respectively

%For the first step, . For the datasets other three types were defined: artificially-generated, benchmark and real-world to evaluate the results at different levels of complexity. Each dataset were clustered by each algorithm ranging the number of clusters $k$ from 1 ($k_{min}$) to 11 ${k_max}$ for a number of trials equals to 30, so that the probabilistic aspects of the algorithms could be evaluated.

\subsection{Data}

Artificially-generated, benchmark, and real-world datasets were used (\tablename~\ref{tab:data}). The artificial datasets were reproduced from the previous work of Tibshirani et al on Gap Statistic~\cite{Tibshirani:2001fj}, in which $5$ datasets were artificially generated according to normally distributed features. For this work, the first dataset was excluded to ensure a fair comparison among the other internal metrics. This dataset arbitrarily assumes that only one group exists and internal metrics such as Silhouette and the CH index are not intrinsically able to retrieve $\hat{k}$ as one. Benchmark datasets have been extracted from the UCI repository~\cite{Dua:2019}, which contains, unlike artificially-generated data, datasets with non-normal statistical distributions, often displaying, for instance, a high skewness.
In this work, a real-world dataset of containing socioeconomic variables at the county-level was obtained from the American Community Survey~\cite{acs}. It includes race, education, and income for each county in the United States. A goodness of fit measurement of a prediction model was used in which the number of heart-failure deaths is predicted based on the following  associated predictors: the total population size, the number of population with diabetes and obesity as well as the percentage of the population with age greater than 65 years. The dataset was obtained from the Centers for Disease Control and Prevention~\cite{cdc}. 
%shows the main characteristics of all data sets.

\begin{table}[ht]
    \centering
    \begin{tabular}{llrrr}
    \toprule
         &            &    $N$ &  $F$ &  $k^*$ \\
    type & dataset &         &         &          \\
    \midrule
    artificial& b &  1000 &    10 &      3 \\
         & c &  1000 &    10 &      4 \\
         & d &  1000 &    10 &      4 \\
         & e &  1000 &    10 &      2 \\
    benchmark & Iris &   150 &     4 &      3 \\
         & Wine &   178 &    13 &      3 \\
         & Wine quality &  1599 &    11 &      6 \\
    real-world & ACS county &  3142 &    21 &      - \\
    \bottomrule
    \end{tabular}
    \caption{\label{tab:data} The characteristics of the datasets.}
\end{table}

\section{Results}

\label{Results}

\subsection{Comparison of Clustering Retrieval among Dataset Types}

\begin{figure}[ht!]
	\centerline{\includegraphics[width=\linewidth]{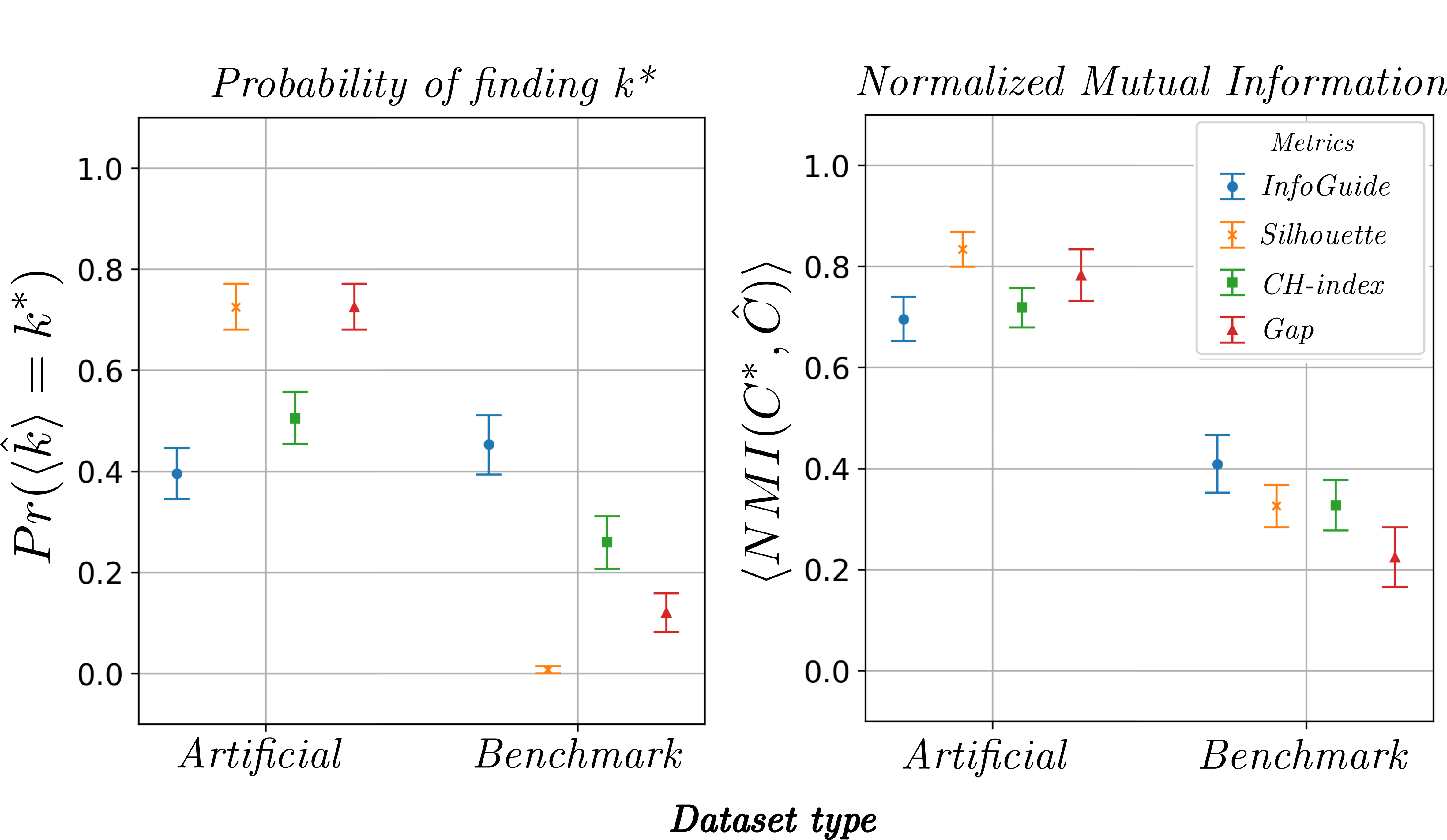}}
	\caption{Comparison of clusters retrieved from artificially-generated and benchmark datasets according to (left) the probability of finding the true number of clusters $k^*$ and (right) the normalized mutual information between the retrieved $\mathcal{\hat{C}}$ and true $C^*$ clusters. 
		% For probability of guessing the truth $k$ the Gap statistics has better results on the artificial data sets while the Select has better results for the benchmark data sets. For NMI estimator most of the metrics are overlapping.
	}
	\label{fig:prob_mutual_types}
\end{figure}

Quality measures of clustering retrieval quantifies to extent to which the retrieved clusters resemble idealized clustering aspects that are often unrealistic when considered the statistical properties underlying the generating process of real-world datasets. Not surprisingly, these measures are largely evaluated over artificially-generated datasets. The clustering retrieval of {\it InfoGuide} was compared against other approaches using both artificially-generated (\figurename~\ref{fig:prob_mutual_types}, left) and benchmark data sets (\figurename~\ref{fig:prob_mutual_types}, right).

Overall, the correct number of clusters is more likely retrieved and a higher information gain is typical obtained in the artificially-generated datasets than in the benchmark datasets. {\it InfoGuide} not only displays the highest information gain in the benchmark datasets but also it displays the smallest decrease in information gain from artificial to benchmark datasets. Though the Silhouette and Gap appears to retrieve superior clusters in the artificial datasets, they retrieve the worst clusters in the benchmark datasets.

\subsection{Comparison of Clustering Retrieval among Algorithms}

\begin{figure}[ht!]
	\centerline{\includegraphics[width=\linewidth]{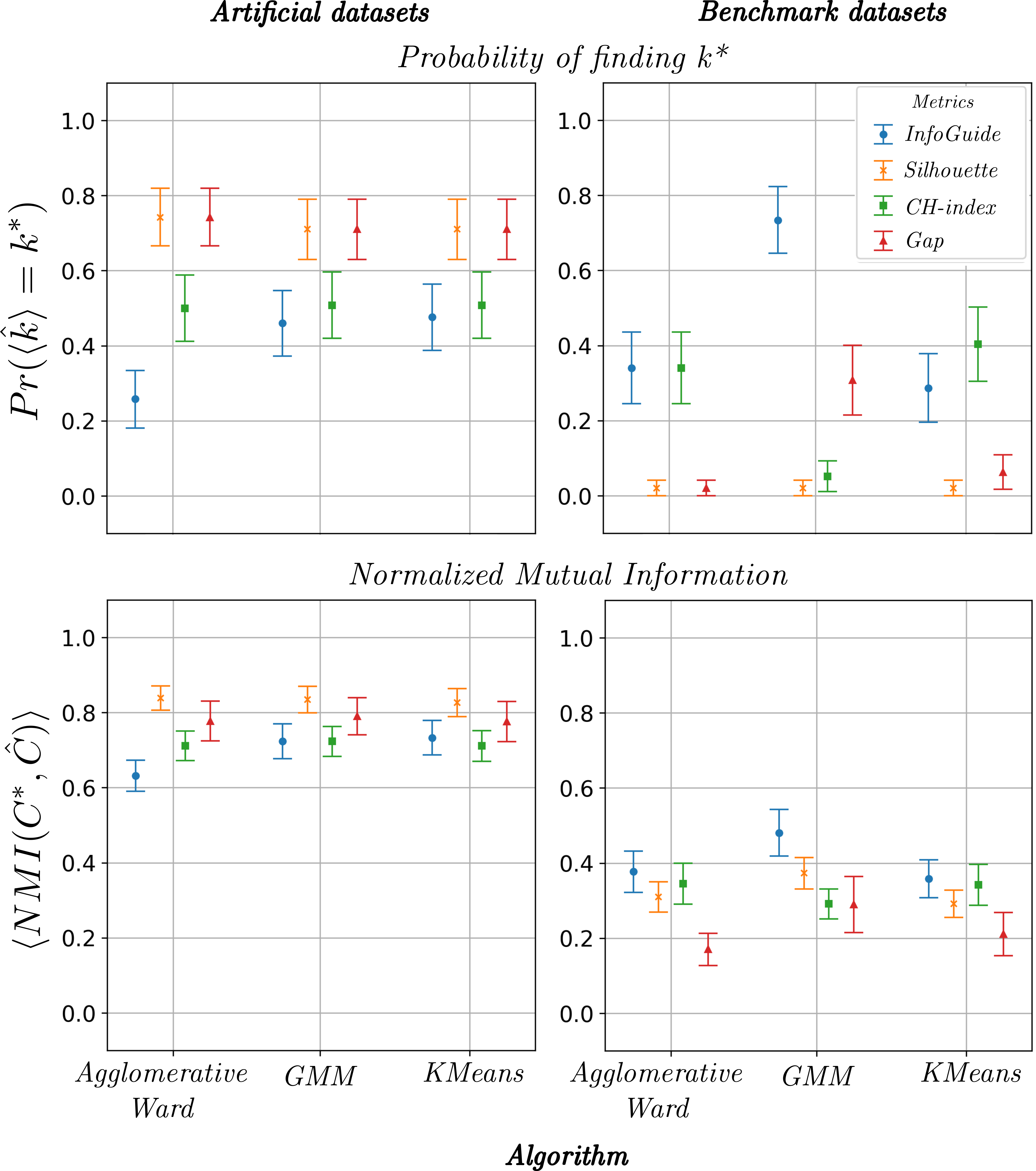}}
	\caption{Comparison of clusters retrieved from artificially-generated (left) and (right) benchmark datasets according to (top) the probability of finding the true number of clusters $k^*$  and (bottom) the normalized mutual information between the retrieved $\mathcal{\hat{C}}$ and true $\mathcal{C}^*$ clusters.
		%	 the Probability of finding the correct $k$ and NMI by algorithm on artificial and benchmark data sets. For artificial data sets the Gap has better results on all algorithms for the probability of finding the correct k, and there is an overlapping between the other metrics for the GMM and K-Means algorithms. For benchmark data sets the Select has the best of all results for the GMM algorithm. For the NMI there is overlapping on the artificial data sets, and for the benchmark data sets the Select has the best of all results for the GMM algorithm.
	}
	\label{fig:prob_mutual_algorithms}
\end{figure}

Generally, each clustering algorithm attempts to retrieve clusters that resemble its idealized aspects of clustering quality defined a priori. Therefore, the clustering retrieval of each algorithm was separately compared according $Pr(\langle k \rangle = k^*)$ and $NMI(C^*, \hat{C})$ using both artificially-generated (\figurename~\ref{fig:prob_mutual_algorithms}, left) and benchmark (\figurename~\ref{fig:prob_mutual_algorithms}, right) datasets.

When the Agglomerative-Ward is used, both Gap and Silhouette tend have retrieved the best clusters in the artificial datasets but the worst clusters in the benchmark datasets. Even {\it InfoGuide} has retrieved the worst clusters when the Agglomerative-Ward is used. Interestingly, Agglomerative-Ward is the only deterministic algorithm and its results may suggest that stochastic components may aid algorithms navigating complex datasets.  

When the algorithms GMM and KMeans were used, each metric retrieved comparable clusters from the artificial datasets according to either $Pr(\langle k \rangle = k^*)$ and $NMI$. Using the benchmark datasets, however, {\it InfoGuide} retrieved the best clusters when the algorithm GMM was used such that when compared to the second-best metric, Gap, {\it InfoGuide} were two times more likely to retrieve the correct number of clusters and also obtained almost two times more information gain.

\subsection{Comparison of Clustering Retrievals in Real-World Datasets}
%\subsection{Heart Failure prediction}

Clustering analysis has been used to find groups in real-world datasets lacking ground truth. To circumvent the absence of ground truth, external validation is commonly used by independently choosing an external dataset of interest that contains metadata associated to all data points within each cluster.  
%Even using one of the most simple models such linear regression,
\begin{figure}
	\centerline{\includegraphics[width=\linewidth]{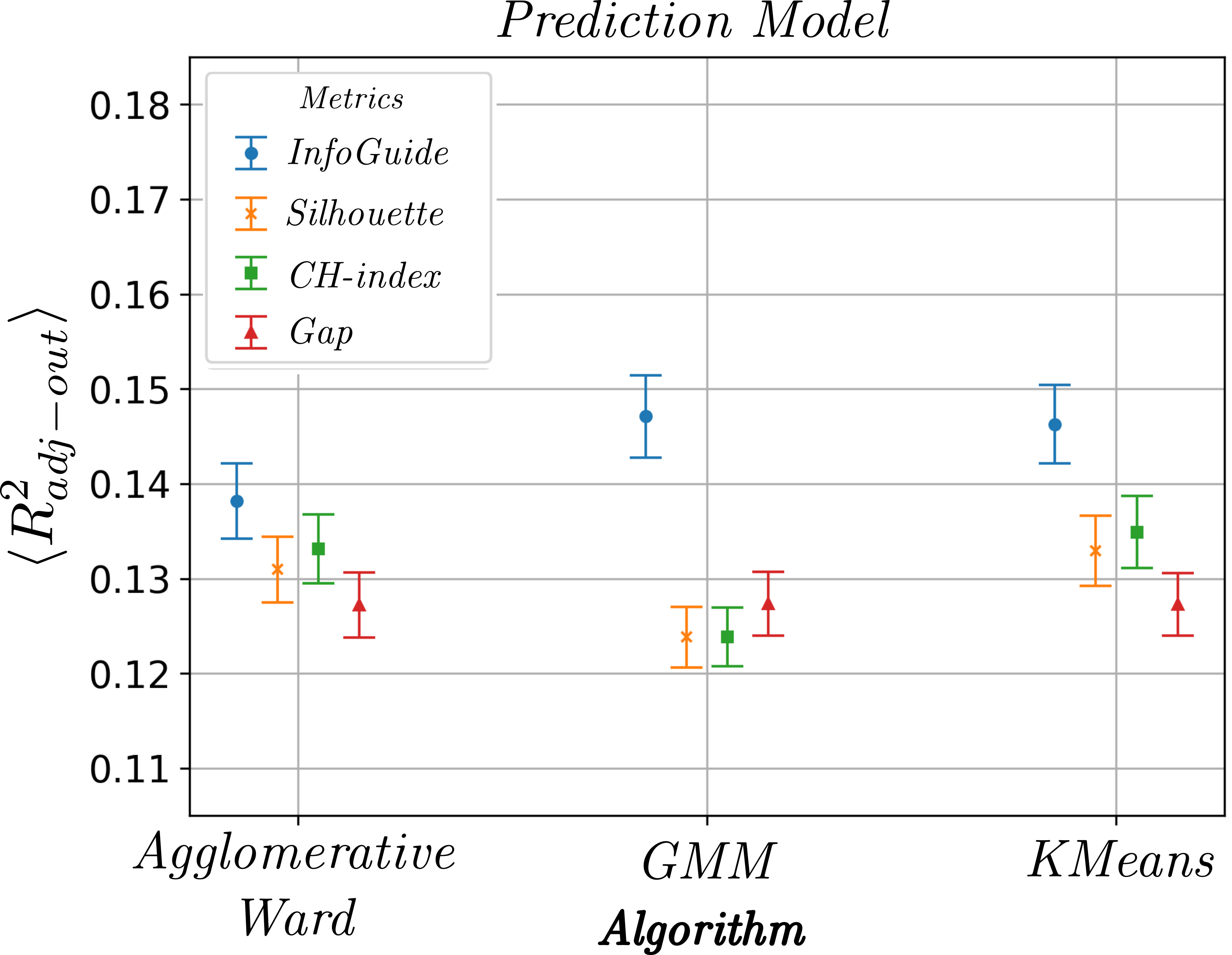}}
	\caption{Results of the Linear regression model for $R^2_{adjusted}$ (out-of-sample) metric. The predict value was the rate of heart failure by county at USA, the clusters found  by each algorithm, guided by each metric, using information about race, income and education (also by county), were used to help the prediction. The clusters found by {\it InfoGuide} added more information to the model in comparison to the other metrics for the GMM and K-Means algorithms.}
	\label{fig:model}
\end{figure}

Overall, the clusters retrieved by $InfoGuide$ obtained the highest adjusted coefficient of determination out-sample $R^2_{adj-out}$ when compared to the other metrics (\figurename~\ref{fig:model}). The clusters retrieved by {\it InfoGuide} were able to explain roughly 3\% more variation of heart failure deaths. Though it is a modest improvement, it can correspond to a total of 100 thousand heart failure deaths incorrectly predicted among the 2.3 million total heart failure deaths in the US.

\section{Conclusions}
\label{Conclusions}
After half-century since the inception of the first clustering algorithm, however, clustering analysis still lacks a more automatic framework for clustering retrieval that is based on internal metrics with less unrealistic assumptions. In this work, we proposed the {\it InfoGuide} method that uses traces of information gain for automatic clustering analysis. 

The results demonstrated that {\it InfoGuide} may be more suitable for retrieving clusters in real-world datasets displaying nontrivial statistical properties. In benchmark and real-world datasets, GMM, which is the algorithm with less strict assumptions, was capable of obtaining the best clustering retrieval. Future works should include a more diverse set of clustering algorithms and datasets from other domains. Despite additional validations, the {\it InfoGuide} method and the idea of using traces of information gain may become a suitable method for automatic clustering analysis.

% \red{Remembering the our contribution} \\ 
% \red{Fuzzy and other data sets} \\

\bibliography{bib}
\bibliographystyle{aaai}
\end{document}